\documentclass[final,5p,times,twocolumn, authoryear]{elsarticle}

\usepackage{amssymb}
\usepackage{booktabs} 
\usepackage{lipsum}
\usepackage{caption}
\usepackage[singlelinecheck=false]{caption}
\usepackage{makecell}
\usepackage{graphicx}
\usepackage{float}
\usepackage{multirow}
\usepackage{hyperref}
\hypersetup{colorlinks,allcolors=black}

\makeatletter
\def\ps@pprintTitle{%
 \let\@oddhead\@empty
 \let\@evenhead\@empty
 \def\@oddfoot{}%
 \let\@evenfoot\@oddfoot}
\makeatother

\begin{document}
\begin{frontmatter}

\title{The Art of Audience Engagement: 
\\LLM-Based Thin-Slicing of Scientific Talks }

\author[a]{Ralf Schmälzle}\fnref{label1}
\fntext[label1]{Corresponding Author. Email: schmaelz@msu.edu}
\affiliation[a]{organization={Department of Communication, Michigan State University},%Department and Organization
            addressline={404 Wilson Rd.}, 
            city={East Lansing},
            postcode={48824}, 
            state={MI},
            country={USA}}
\author[a]{Sue Lim}
\author[a]{Yuetong Du}
\author[a]{Gary Bente}

\begin{abstract}
This paper examines the thin-slicing approach – the ability to make accurate judgments based on minimal information – in the context of scientific presentations. Drawing on research from nonverbal communication and personality psychology, we show that brief excerpts (thin slices) reliably predict overall presentation quality. Using a novel corpus of over one hundred real-life science talks, we employ Large Language Models (LLMs) to evaluate transcripts of full presentations and their thin slices. By correlating LLM-based evaluations of short excerpts with full-talk assessments, we determine how much information is needed for accurate predictions. Our results demonstrate that LLM-based evaluations align closely with human ratings, proving their validity, reliability, and efficiency. Critically, even very short excerpts (\textless{} 10\% of a talk) strongly predict overall evaluations. This suggests that the first moments of a presentation convey relevant information that is used in quality evaluations and can shape lasting impressions. The findings are robust across different LLMs and prompting strategies. This work extends thin-slicing research to public speaking and connects theories of impression formation to LLMs and current research on AI communication. We discuss implications for communication and social cognition research on message reception. Lastly, we suggest an LLM-based thin-slicing framework as a scalable feedback tool to enhance human communication.

\end{abstract}


\end{frontmatter}

\section*{Introduction}
Humans instinctively form rapid impressions of others based on minimal cues. While research has demonstrated that such thin slices of social behavior \citep{Ambady1993Half} are surprisingly accurate across many domains, their applicability to evaluating complex scientific presentations remains less explored. Conventional wisdom and the popular science literature on public speaking suggests that we often judge a speaker's competence within seconds of them taking the stage \citep{Ailes2012You}, but empirical research remains scarce. In this study, we investigate whether thin slices of scientific presentations – just like the talks given at academic conferences – can reliably predict the presentations’ overall effectiveness. We introduce a novel dataset and new methods based on Large Language Models (LLMs) to analyze the predictive power of these fleeting initial moments.

This paper is structured as follows. First, we introduce the topic of thin-slicing and discuss how it has been studied in nonverbal communication and social perception and cognition research. Then we zoom in on the topic of public speaking and discuss how research on social impression formation and thin-slicing relates to this domain. Third, we discuss how the advent of LLMs offers new ways to study thin-slicing productively and with unprecedented efficiency. We then introduce the current study and its hypotheses, which focus on the evaluation of thin-sliced speech transcripts of a large corpus of science communication talks, followed by specific methods, results, and discussion.

\begin{figure*}[hbt!]
        \centering
	\includegraphics[width=0.75\textwidth]{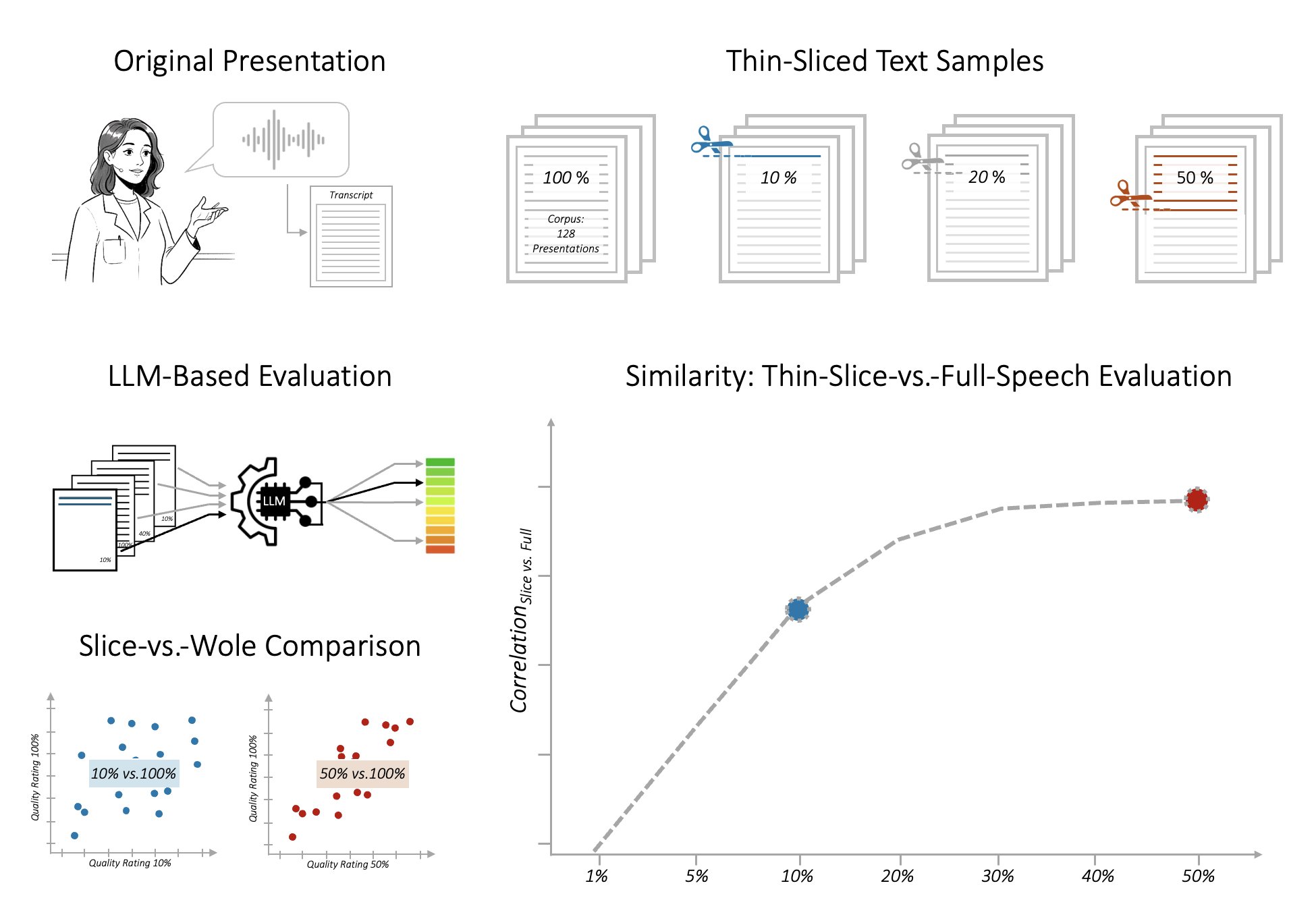}	
	\caption{Logic of Thin-Slices Evaluation of Public Speech Performance in the Context of Science Communication. Recordings of real-life talks about science topics are transcribed to text. Next, each transcript is thin-sliced into excerpts containing either the full speech text, or slices corresponding to 1\%, 5\%, 10\%, etc. These slices are then submitted to LLM for quality assessment, leading to a table with ratings for all speeches and across all slices. Ratings are collected independently (no memory in the LLM) and evaluated via different prompts and multiple LLMs. Finally, we compare evaluations across slices to examine how much of the speech needs to be processed until stable quality predictions can be made. } 
	\label{fig_mom0}%
\end{figure*}
 
\subsection*{Thin Slices of Social Behavior }
A large body of research from social psychology and nonverbal communication suggests that we can often glean surprisingly accurate social information from only short glimpses into others’ observable behaviors \citep{Ambady1992Thin, Uleman2022Differences}. A classic example of social perception based on such thin slices comes from classroom teaching: \citet{Ambady1992Thin, Ambady1993Half} found that observers could predict a teacher’s end-of-semester evaluations after watching only a brief, silent video clip of that teacher’s classroom behavior. In their study, even 30 seconds of nonverbal interaction (without audio/verbal content) provided enough information for strangers to assess teaching effectiveness, forecasting evaluations given by the teacher’s actual students months later. 

This and similar studies launched the idea that rapid, minimal exposure to a person’s behavior can reveal stable qualities \citep{Ambady2010Perils}. In other words, relevant information (i) must be expressed and (ii) can be extracted from only a thin slice of the whole. Thus, the so-called thin-slicing paradigm typically involves presenting observers with brief excerpts of behavior – often just seconds-long videos of nonverbal behavior or even still pictures – and asking them to make judgments. Then, by correlating ratings that were made based on thin slices with ratings based on exposure to the whole interaction, it can be established how much information is needed to arrive at a stable judgment that is predictive of the whole. Meta-analyses show that across a broad range of social domains \citep{Murphy2021Capturing, Slepian2014ThinSlice}, thin slices can predict consequential outcomes like interpersonal warmth, personality characteristics, physician competence, relationship quality, or the outcome of legal proceedings \citep{Carcone2015Comparing, Houser2007Predicting, Krumhansl2010Plink, Nguyen2015Would, Parrott2015Thin}. 

While most existing thin-slicing literature focused on nonverbal behavior, this paradigm is also applicable to paraverbal and verbal domains \citep{Hall2021ThinSlice, Slepian2014ThinSlice}. For example, we can make rapid judgments from a voice or even a written email, like inferring a sender’s gender and age during a phone call, or their emotional state or personality from their writing. Along similar lines, we may be able to infer a person’s competence, confidence, or enthusiasm soon after they start speaking \citep{DErrico2013Perception, Gheorghiu2020ThinSlice, Rosenberg2009Charisma}.

\subsection*{Translating the Thin-Slicing Approach to the Public Speaking Domain}
Evidently, the public speaking situation overlaps with that of classroom teaching, the domain in which thin-slicing research originated from. Both involve presentations by a speaker to an audience, i.e., a one-to-many form of communication that blends the interpersonal and mass communication domains \citep{Berger2010Handbook}. Although the thin-slicing approach stems from social psychology and education, almost all application contexts are communicative in nature (e.g., relationships, business, health, or legal interactions), focusing on the expression of social signals by senders and their perception by recipients. Thus, thin-slicing is very applicable to communication research in general and to public speaking in particular. This all suggests that thin-slicing research is highly relevant to public speaking. Indeed, in the popular literature on public speaking education, there is a widely cited notion of a “seven-second rule,” suggesting that listeners decide within the first few moments of a talk whether the speaker is worth their attention \citep{Ailes2012You}. Although closer inspection shows that this rule is based largely on anecdotal data, it is widely assumed and taught that a strong start matters greatly in presentations 
\citep{Hey2024Mastering, Lucas2020Art}. This aligns directly with thin-slicing research as well as work on first impressions more broadly \citep{Todorov2017Face}.

Some limited research has connected these domains, but they mostly address informal or non-scientific content \citep{Chollet2017Assessing, Cullen2018Perception, Feng2019Proof, Ismail2016ThinSlices}. For instance, in an observational study of a meeting, \citet{Ewers2018Do} found that the degree of dullness of a talk after 4 minutes predicted whether the speaker would “ponder on”, i.e., go into overtime. Another related study by \citep{Cullen2017ThinSlicing} used the thin-slicing approach to TED talks by applying machine learning feature extraction techniques to visual and acoustic parameters. While this work is directly relevant to the current study, its focus on preselected TED talks, which tend to be optimized for entertaining popular audiences, sets it apart from our focus on scientific presentations. Other work by Chollet and Scherer \citep{Chollet2017Assessing}  also directly connects the thin-slicing literature to public speaking. In a study of 45 speakers giving informal impromptu presentations about the city of Los Angeles and a beauty product, \citep{Chollet2017Assessing} found that automatically assessed audio-visual features forecasted speech evaluations. 

Perhaps the two most directly related prior studies are \citet{Feng2019Proof} and \citet{Ismail2016ThinSlices}.  Ismail presents an elaborate proposal on how thin-slicing could be applied to public speaking evaluation, but without empirical data. Feng and colleagues, who are part of the Educational Testing Service corporation, conducted a thin-slicing study with 17 speakers who were recruited via the Toastmasters organization and gave speeches about different pre-assigned topics. However, their study focused largely on the visual and nonverbal delivery factors, assessed via video-based thin-slicing and human ratings. Although they discuss speech content quality, it was not directly examined via thin-slicing. Overall, there is a need for larger and more systematic investigations of science communication quality leveraging the thin-slicing approach. 

\subsection*{Potential of Large-Language Models to Enable Thin-Slicing Studies of Public Speaking}
Even though the thin-slicing paradigm originated from an inherently communicative situation (classroom teaching), the domains of thin-slicing research in social cognition and public speaking training and assessment have remained surprisingly distant. This gap between the thin-slicing literature and the literature on public speaking competency appears partly due to the practical challenges of studying realistic public speaking performances, which is labor-intensive and requires a large corpus of real speeches and many raters . For example, Ambady and Rosenthal's  work involved only 9 raters who had to watch and manually code all 39 video clips \citep{Ambady1993Half}. Feng et al. highlighted the enormous burden these tasks place on raters \citep{Feng2019Proof}. And the same challenges apply to the large body of research that uses human coders to study topics like social perception and impression formation \citep{Grahe1999Importance, Schmalzle2019Visual, Willis2006First, Wallbott1986Cues}, as well as more focused investigations of speaker ability, charisma, and similar topics \citep{Gheorghiu2020ThinSlice, Rosenberg2009Charisma, Cullen2018Perception}. In sum, the inherent challenges are clear: high expense and rater wear-out must be balanced with efficiency representativeness, and other constraints.

LLMs pave the way for closing the gap between thin-slicing research and the perception and evaluation of speech performances. LLMs are advanced artificial intelligence systems trained on vast amounts of text data, enabling them to generate human-like text and perform various language-based tasks \citep{Tunstall2022Natural}. This development in LLMs have majorly shifted how we interact with and utilize AI, with models transforming industries and reshaping the future of communication and information processing \citep{Bishop2024Deep, Mitchell2019Artificial}. 
LLMs offer a potential remedy for the challenges related to human raters discussed above \citep{Argyle2023OutOfOne, Calderon2025Alternative,Gilardi2023ChatGPT, Markowitz2024Generative}. LLMs can perform complex tasks, including evaluating speech transcripts in terms of basic tasks like word counts, and also higher-level impressions about social characteristics \citep{Bubeck2023Sparks, Dillion2023CanAI}. While it remains an empirical question whether LLM-based evaluations align with those made by humans, communication scholars are well-equipped to investigate it \citep{Krippendorff2004Content, Neuendorf2017Content, Riff2014Analyzing, Shrout1979Intraclass}. Thus, by demonstrating correlations between human and LLM evaluations of the same speeches, we can validate the use of LLMs for evaluating public speeches, which could scale up thin-slicing research far beyond what was previously feasible and open new avenues of investigation for understanding and improving communication.

\subsection*{The Current Study}
Building on the research streams summarized above, this study leverages LLMs to examine how much information is needed to predict science presentations’ overall rhetorical quality. We focus on science presentations because they constitute a setting where effective communication is critical \citep{Doumont2009Trees, Fischer2024Affect, Gu2007Ten, Hey2024Mastering}. Science presentations, like the typical talks given at conferences, require engaging an audience quickly on complex topics, keeping them attentive over a sustained period, and presenting information in such a way that it is comprehensible to audiences who are intellectually able and motivated to learn, but may not be familiar with the minutiae of the research. By testing the predictive power of early impressions, we thus aim to the literature on impression formation and practical public speaking assessment. 

 To conduct the study, we first collected a large corpus of over 100 science presentations. Each presentation lasted between 8-12 minutes and consisted of actual research-based content that the speakers prepared (e.g., talks for the upcoming conference or job talk). We then used LLMs to rate the quality of presentation transcriptions. For each transcript, the LLMs rated thin-sliced subsets with slice lengths varying parametrically (e.g. the first 50\% of the speech, 40\%, 30\%, 20\%, 10\%, 5\%, 1\%) as well as the entire speech. We validated this LLM-based method using human raters.
 
Through this LLM-based thin-slicing approach, we address the following hypothesis and research questions. Based on the literature on thin-slicing summarized in the introduction, we predict that the quality ratings of thin-sliced subset of the presentations’ text transcripts would positively correlate with the quality ratings of the entire transcript (i.e., show a thin-slicing effect; H1). 

Next, we examined how much content is required in the thin slice to best predict the quality ratings of the overall speech (RQ1). We estimated that less than 20 percent of a speech (about 3 minutes) should be enough \citep{Ewers2018Do} and aimed to pinpoint a more precise moment when additional information no longer becomes relevant. Finally, we explored whether the findings for H1 and RQ1 would differ by the specific LLM used and instructions provided to the LLMs (RQ2). 

To our knowledge, no other study to-date has applied LLMs to evaluate public speaking quality; thus, our LLM-based thin-slicing approach makes significant contributions to the communication discipline as well as social cognition research. If the AI’s thin-slice ratings correlate strongly with full-speech outcomes, then that could point toward practical tools for rapid, automated feedback on presentations. Across the learning sciences, it is clear that feedback is a key ingredient of improvement \citep{Domjan2020Principles, OECD2010Educational, Skinner1961Teaching, Silver2021Reward}. 

However, many existing feedback systems for public speaking do not provide real-time information (e.g., content on slide 2 could be articulated more clearly) and generates arbitrary qualitative or composite scores that limit speaker improvement. LLM-based feedback systems can potentially address these limitations, giving speakers a quick ability to change course. Furthermore, comparing LLM and human ratings can also give us theoretical insights into how closely these models mirror human perceptions of rhetorical quality or perceived effectiveness.

\section*{Methods}
\subsection*{Public Speaking Corpus}
We assembled a corpus of 160 public science presentations for analysis. Eighty members of the scientific community (graduate students and faculty) presented two separate research-based talks about the area of their expertise. Thus, the speakers had ample time to prepare their talks for professional science audiences and had a personal interest in the talks being of high quality. The talks were presented in front of a large audience in a virtual reality (VR) environment that resembled a professional venue (conference-style hotel room). 

The talks were 8-12 minutes long and were recorded. Then we transcribed the presentations using OpenAI’s Whisper model and manually checked the transcriptions for errors. We also screened the transcripts for quality and removed talks with incomplete recordings or poor audio. After this filtering, 128 speeches remained in the corpus. 

Overall, this corpus amounted to over 100.000 seconds and almost a quarter million of spoken words – about an entire conference days’ worth of public speeches. By working with text transcripts, we aimed to provide a consistent input to the language models and to enable human raters to evaluate content without being influenced by visual or audio cues.

\subsection*{LLM-Based Thin-slicing Procedure }
All analyses were conducted in Python, and we fully document the analysis pipeline online. First, the text transcriptions were loaded and sliced (i.e., subsampled) into the first 1\%, 5\%, 10\%, 20\%, 30\%, 40\%, 50\%, 75\%, and 100\% of the speech. We also ran the same analyses using fixed numbers of words (because the percentages can contain different numbers of words). However, the conclusions hold, and thus we only report the percentage-based results here in the main text (see Supplementary Materials for additional details).

\subsection*{LLM-Based Speech Evaluation Procedure: Models and Prompting Strategies }
We deployed two advanced language models, GPT-4o-mini and Gemini Flash 1.5, as AI evaluators. These models were chosen for their strong language understanding capabilities (Omar et al., 2025), which would enable them to judge coherence, clarity, engagement, and other qualities of the speeches. GPT-4o-mini and Gemini Flash 1.5 each evaluated all 128 speeches in all slices, yielding a set of AI-generated scores for every full speech and every excerpt. 

We experimented with five different prompt formulations for each model to ensure robustness of the AI’s responses (See Supplementary Materials for the exact prompts). By using multiple prompts, we checked that the LLMs’ ratings were not overly sensitive to prompt phrasing or context. In total, this approach yielded 11520 ratings, and the entire submission of speeches and retrieving the ratings took about 1 hour. 

\subsection*{Human Evaluation and LLM Validation }
Because the use of LLMs in a way that mimics human raters is still evolving, we also conducted a human rating study to validate that LLM’s ratings capture meaningful variation that can be perceived by humans. To this end, we recruited a group of 60 human raters (\textit{mean\textsubscript{age}} = 38.2, sd = 10.7, 24 self-identified males) to read through the speeches and rate the rhetorical quality. The study was approved by the local IRB and all raters provided informed consent and received \$4 for their evaluations, which took about 20 minutes. 

Procedures were kept as parallel as possible as for the LLM-prompt. Because internal tests revealed that raters would have difficulties reading the entire speeches, we decided to evaluate only the 20\% version. This amounted to about a half page to a page of text, which is feasible in terms of its attentional demands. The 60 raters were split into two groups, and each evaluated a sample of 12 speeches drawn from the corpus of 128 speeches. A total of 24 speeches were evaluated.

\subsection*{Statistical Analysis Methods}
Once human ratings were collected, we examined interrater-agreement among the human raters based on intra-class methodology \citep{Shrout1979Intraclass}. In parallel, we also assessed the agreement between different LLM models and prompting strategies. Next, we computed Pearson correlations between group-averaged speech quality ratings and the LLM’s ratings for each speech \citep{Pearson1895Note}. 

Lastly, following prior work on thin slice judgments, we computed Pearson correlations between the ratings for each slice and the rating for the entire speech. A high correlation between the slice and the entire speech suggests that the relevant information can be successfully extracted already within a much shorter slice. Our goal was to measure the threshold where this correlation reaches significance.

\hphantom{<text>}  

\hphantom{<text>}    

\begin{figure*}[hbt!]
        \centering
	\includegraphics[width=0.65\textwidth]{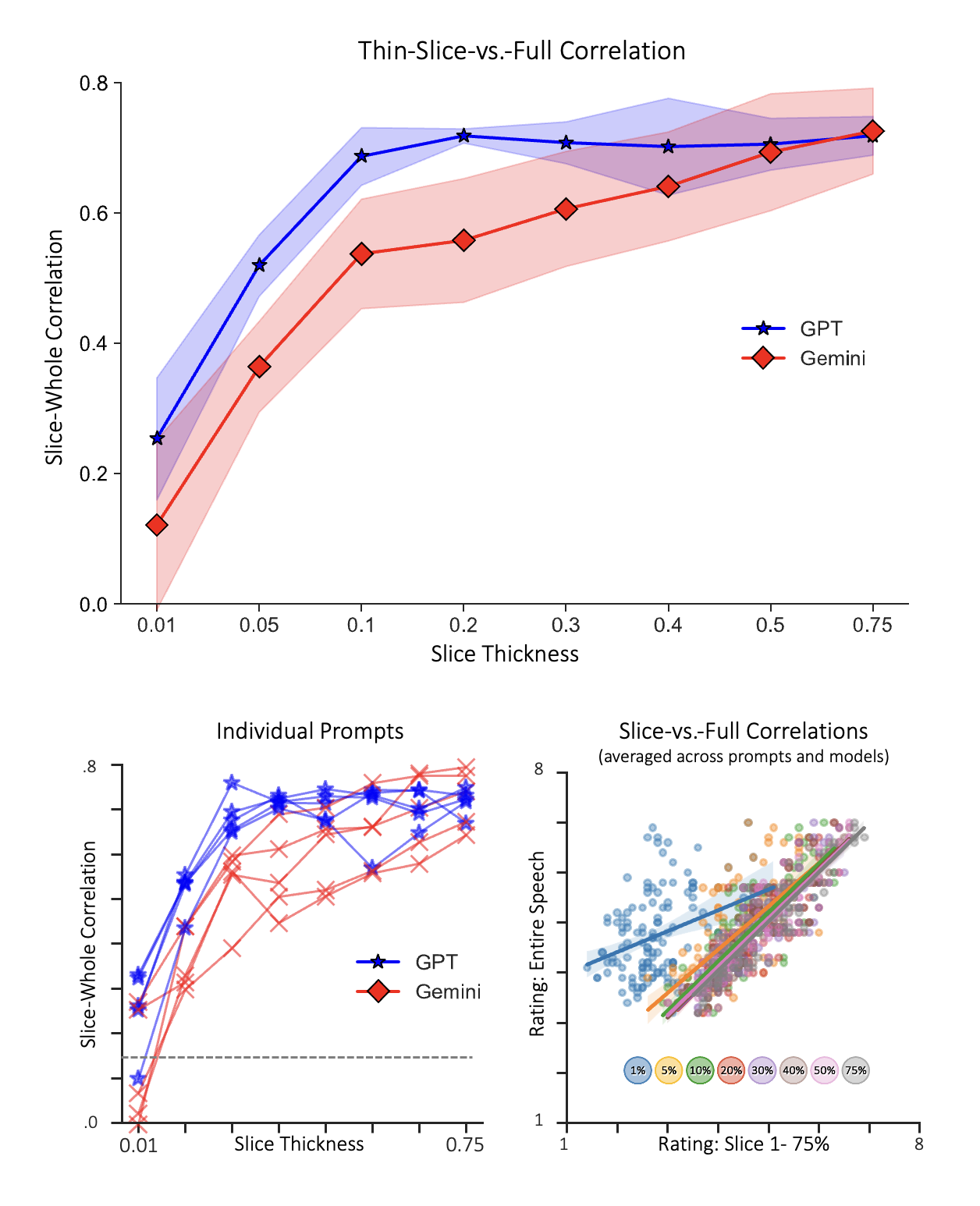}	
	\caption{Thin-Slice to Full-Speech (part-to-whole) Correlations for both LLMs. Shaded corridors illustrate the variability across the five different prompts. Bottom panels: Left: Individual-prompt results for OpenAI’s GPT (blue) and Google’s Gemini (red) models. As can be seen, the same general pattern is present regardless of model family or prompt wording. Right: Scatter plots for all 128 speeches. As slice thickness increases, the predictions become progressively more aligned with the evaluation for the entire speech.
} 
\label{fig_mom0}%
\end{figure*}

\section*{Results}

\subsection*{Interrater Agreement and Convergence of LLM and Human Ratings}
Starting first from the human ratings (60 raters evaluating 24 speeches, split into two samples), we find that the human raters exhibited high consistency in their speech evaluations, as demonstrated by high-intra-class correlations \textit{ICC\textsubscript{2,1}} = 0.92 and 0.86 \citep{Shrout1979Intraclass}. This demonstrates that raters demonstrated high levels of agreement about speeches’ perceived quality rankings; the high consistency also underscores that the group-averaged quality evaluations per speech are highly reliable and no further benefits would accrue from using additional raters) \citep{Kelley1925Applicability, Kim2019Reliable, Kraemer1992How}.

Next, we conducted a parallel stream of analyses to assess agreement among different ways to elicit LLM evaluations (RQ2). In other words, we applied the same procedures as conducted for the human ratings to the data obtained for the different LLM models and the five prompts – essentially treating the model/prompt-instances as if they were 10 raters. This analysis yielded a high inter-model/prompt-agreement, \textit{ICC} = 0.93 – similar in magnitude as observed for the human raters , suggesting that the evaluations did not differ significantly by the specific models and prompts (see Supplementary Figure 1 for details), but the specific prompt phrasing barely influenced ratings. Overall, this suggests that our findings are not an artifact of a peculiar prompt or an idiosyncrasy of one AI system. Instead, they reflect a stable AI-based assessment of speech quality that emerges regardless of how we queried the models. 

% Use table* instead of table
\begin{table*}[htbp] % Note the asterisk (*)
\centering
\caption{Comparison of Model Performance across Slice Thickness.}
\label{tab:slice_thickness_span}

% --- The rest of your table code is exactly the same ---
\begin{tabular}{llcccccccc}
\hline\hline
\multicolumn{2}{c}{} & \multicolumn{8}{c}{\textbf{Slice Thickness}} \\
\cline{3-10}
\textbf{Model} & \textbf{Prompt} & \textbf{0.01} & \textbf{0.05} & \textbf{0.1} & \textbf{0.2} & \textbf{0.3} & \textbf{0.4} & \textbf{0.5} & \textbf{0.75} \\
\hline
\multirow{5}{*}{GPT} & Prompt\#1 & 0.32 & 0.54 & 0.65 & 0.72 & 0.71 & 0.74 & 0.74 & 0.67 \\
                     & Prompt\#2 & 0.26 & 0.53 & 0.68 & 0.73 & 0.67 & 0.74 & 0.74 & 0.73 \\
                     & Prompt\#3 & 0.33 & 0.54 & 0.69 & 0.72 & 0.75 & 0.73 & 0.70 & 0.75 \\
                     & Prompt\#4 & 0.25 & 0.55 & 0.76 & 0.72 & 0.73 & 0.73 & 0.69 & 0.72 \\
                     & Prompt\#5 & 0.10 & 0.44 & 0.65 & 0.70 & 0.68 & 0.57 & 0.65 & 0.72 \\
\hline
\multirow{5}{*}{Gemini} & Prompt\#1 & 0.02 & 0.33 & 0.55 & 0.54 & 0.64 & 0.66 & 0.70 & 0.74 \\
                        & Prompt\#2 & 0.27 & 0.44 & 0.60 & 0.61 & 0.66 & 0.66 & 0.78 & 0.80 \\
                        & Prompt\#3 & 0.00 & 0.44 & 0.58 & 0.69 & 0.70 & 0.76 & 0.78 & 0.78 \\
                        & Prompt\#4 & 0.25 & 0.31 & 0.56 & 0.45 & 0.51 & 0.56 & 0.58 & 0.64 \\
                        & Prompt\#5 & 0.07 & 0.30 & 0.39 & 0.50 & 0.52 & 0.56 & 0.63 & 0.67 \\
\hline
\multicolumn{2}{l}{\textbf{Average}} & \textbf{0.19} & \textbf{0.44} & \textbf{0.61} & \textbf{0.64} & \textbf{0.66} & \textbf{0.67} & \textbf{0.70} & \textbf{0.72} \\
\hline\hline
\end{tabular}
% --- End of your table code ---

\end{table*} % Note the asterisk (*)

Having established that human raters as well as different LLM models with specific prompts each produce convergent ratings among each referent group, we proceeded to examine whether human and LLM-based evaluations also converge. To this end, we correlated the group-averaged speech evaluations from the human sample with the corresponding ratings from the LLM-based speech evaluations. We find that the correlation amounts to \textit{r\textsubscript{human-rating-vs.-LLM-rating}} = 0.69, which is highly significant (\textit{t(22)} = 4.47, \textit{p} \verb|<| 0.0001). This shows that regardless of whether speeches are evaluated by humans or by LLMs, both evaluation modes yield very similar conclusions about which speeches are considered high vs. low quality. This again underscores the promise of LLM-based evaluations in terms of validity, but with much higher efficiency (see Supplementary Figure S1).

\subsection*{Main Analysis: Thin Slice Correlations}
The central question of our study was whether a thin slice of a public speech, like the first 10\%, can predict the rhetorical quality of the entire speech (H1 and RQ1). Our results suggest that it can. Figure 2 plots the strength of the correlation between each part (slice) and the entire speech. As can be seen, correlations rise quickly across progressively thicker slices and converge at around 0.6/0.7. This effect is visible for both LLMs – whether from the Gemini or GPT4o-family. Furthermore, already at 10\% of the entire speech, the plateau is basically reached, suggesting that from this point onwards, additional incoming speech content does not make much of a difference in terms of the overall evaluation.

Interestingly, even very thin slices, such as 5\% or even 1\% of the entire speech, show positive correlations. With the sample size of 128 speeches, the chance-level (i.e., \textit{a} = 0.05) lies at \textit{r} = 0.145. Even at the 1\%-slice, most correlations (4 out of 5 for the GPT-based prompts and 2 out of 5 for the Gemini model) are above this threshold, and the threshold is passed for all (10 out of 10) model/prompt-configurations at the 5\% slice. In the current sample, these slices correspond to just 15 (for 1\%) and 60 (for 5\%) words. 

\section*{Discussion}
This study examined whether early impressions of public science presentations can predict the presentation’s overall evaluation. We tested the potential of language models can reliably emulate speech evaluations in this context. Our findings provide strong support for the thin-slicing effect, suggesting that a brief exposure to a presentations’ transcript contains information that enables predictions of its overall rank in terms of rhetorical quality. Furthermore, we find that LLMs are accurate and efficient.

The results demonstrate that a thin slice of a presentation’s written transcript allows forecasting its overall quality. In fact, even the very first few sentences contain predictive information that enabled correlations between 0.3 and above (see Figure 2). A plateau effect emerged starting at about 10\% of the speech, suggesting that relevant information has been expressed at this point and evaluations do not change much from there on. Notably, even extremely brief excerpts—just 5\% or even 1\% of the full speech—exhibit positive correlations. In this dataset, these slices equate to roughly 15 words (1\%) and 60 words (5\%). Considering typical public speaking rates of 100–150 words per minute, this aligns remarkably well with the “7-second rule” \citep{Ailes2012You}.

This main result was stable across different analysis methods and was seen similarly using different LLM models and prompt wordings. Moreover, it is important to highlight that the demonstrated correspondence between human evaluations and LLM-based evaluations validates the use of the latter in the first place. If there was no strong correlation between human and LLM-evaluations, or if the correlation was only moderate, then one could question the validity of LLMs. However, given that LLMs’ evaluations of the speeches converge with those of human raters, we can confidently use LLMs and thereby greatly increase the efficiency of thin-slicing studies, which are very time-consuming, costly, and taxing. In fact, the core of our LLM-based speech evaluation consists of a few lines of code in which a for-loop prompts the API of GPT4 and Gemini, respectively. With this pipeline, the independent evaluation of all 128 speeches across 2 models, 5 different prompts, and for the entire speech as well as 8 sub-slices (1 - 75\%) took less than half an hour, costing less than \$5. 

Compared to the ca. \$150 we paid for the human evaluation study, which only comprised a small fraction of the volume (i.e., only one task instruction/prompt, one slice, and 24 speeches), it is easy to see the superiority of the LLM-based assessment in terms of cost-effectiveness and scalability. This is not to say, however, that human evaluations are no longer necessary: it still needs to be demonstrated that LLM-based evaluations are consistent and converge with human impressions. But once that is established, as in the current context of evaluating speech transcripts, the pendulum swings clearly in favor of using LLMs \citep{Argyle2023OutOfOne, Calderon2025Alternative, Dillion2023CanAI, Eger2025Transforming, Gilardi2023ChatGPT}. 

\subsection*{Theoretical Implications and Practical Applications}
In the following section, we first discuss the theoretical implications and connections of the current findings with the communication science literature and then point out practical applications. 

The thin slices/first impressions literature provides empirical backing for Uncertainty Reduction Theory \citep{Berger1975Some}. In particular, one of uncertainty reduction theory’s core ideas is that people use limited salient information to make quick judgments to guide social interactions. To our knowledge, these connections have not been well articulated, and thus, the thin-slicing/first impressions literatures in psychology and uncertainty reduction theory in communication have evolved somewhat in parallel. However, by focusing on the initial encounter situation in which a new speaker presents themselves to an audience, it is natural to see how the two bodies of research converge: At the beginning of a talk, there is necessarily high uncertainty about what is going to happen next, what the talk will be about, and whether the speaker can get the point across. But this uncertainty is progressively resolved as the talk unfolds and audience members form impressions. The accuracy of these snap impressions has long been a debate in the social psychology literature \citep{Jussim2017Precis}, but in the case of rhetorical quality judgments, the current results suggest that it can indeed be quickly sensed how good a speaker/speech is.

Another relevant body of work, again unconnected to thin-slicing research, can be found in classical newspaper readership studies in journalism and mass communication. For example, \citep{Schramm1947Measuring} showed that most readers of long-form newspaper articles stopped reading early on, as if they lost attention or found the text too long and dry. In fact, studies of article reading depths resulted in new media formats, such as the USA today newspaper with its brief articles. Nowadays, similar evolutionary developments seem to unfold with online texts \citep{Berger2023What}, simple choices like whether to read an article based on a headline \citep{Scholz2017Neural}, but also with short video formats like TikTok and YouTube Shorts. Critically, the point is that readers make snap judgments about whether the content is interesting and whether it is worth to keep reading. Relatedly, there is also renewed interest in people’s sequential media choices, i.e., how people choose between different songs, videos, books, and so on \citep{Gong2023Media}. 

The work presented here connects these lines of inquiry insofar as it focuses on the choice within a given message (like a speech, but also a book, song, TV show, or newspaper article), i.e., how people make decisions implicitly about staying engaged. To avoid misunderstanding though, we have not yet studied here whether real audiences would “tune out” of some low-quality speeches after 10\%, but recent work in neuroscience of audience response measurement suggests that this could be feasible \citep{Schmalzle2015Engaged, Schmalzle2022Theory}.

The current results are not only theoretically interesting regarding the nature of public speaking and how a speaker’s skills are expressed as the speech unfolds, but they can also improve communication training and practice: By demonstrating the effectiveness of thin-slicing and the feasibility of LLMs for speech transcript analysis, we offer a pathway toward automated and scalable feedback and augmentation tools for speakers. Especially with automated public speaking training, such tools could offer valuable, immediate, and actionable feedback \citep{Forghani2024Evaluating, VallsRates2023Encouraging}. For example, even within standard software tools like Microsoft PowerPoint, there is already a tool called “Speaker Coach” \citep{Microsoft2025Rehearse}, which allows speakers to rehearse their slide shows and provides basic feedback about speech rate and overused filler phrases. However, this tool is very basic and does not give feedback about the content or organization of the presentation itself. These are areas where LLMs could help a lot to improve the speakers’ notes, making them clearer, easier to understand, and ultimately more effective \citep{Shulman2024Reading}. 

In sum, the work presented here about LLM’s capabilities to swiftly detect early warning signs of a talk that might be at risk of losing the audience could empower scientists and other professionals to refine their communication skills. This could lead to more effective dissemination of complex information to audiences. Given that science communication is crucial for public understanding of science, this could have great benefits. Moreover, the methodological framework developed here could be applied to other communication domains that build on public presentation skills, such as education, business, and politics.

\subsection*{Strengths, Limitations, and Avenues for Future Research}
This study's strengths include its novel application of thin-slicing to public speaking, specifically focusing on the verbal communication channel. Also, the use of a large and high-quality corpus of science communication talks, and the exploration of multiple LLMs and prompts are positives. However, the study is limited by its focus on a specific type of communication (science talks) and a specific population of speakers (academics), although this homogeneity and expertise could also be viewed as an asset. 

Finally, the reliance on transcripts means that nonverbal cues, known to influence communication, were not directly analyzed in this study. Integrating the verbal and nonverbal cues – as well as analyzing each channel’s contribution separately – are valuable steps that we are working on next \citep{Wolfe2014Multichannel}. Clearly, it needs to be kept in mind that the speeches were originally delivered verbally but then transcribed into written text. Spoken, conversational language differs from written expression, the former being more informal. This does not, however, challenge our findings because even if the LLM-based evaluations would punish against informal speech, this would equally apply to all speeches, keeping their relative ranking intact. But we do note that the act of transcribing itself could lead to a loss of information and that we noted some cases where we had to correct, e.g., transcription errors based on foreign speakers’ accents, and other factors. Also, the influence of paralinguistic factors, such as “uhs”, “ums”, mumbling, or even stuttering is a topic worth mentioning. In our transcription process, we purposely kept filler words and occasional word repetitions in the corpus as these dysfluencies do contain diagnostic evidence; but we corrected mumbling, thus making the transcripts clearer than the spoken speech would likely have been perceived. 

But this all points to the broader distinction of the domains of speech content/organization vs. speech delivery, which are core to public speaking skills \citep{Aristotle2013Poetics, Lucas2020Art}. In sum, while the current work demonstrates the promise of LLM-based thin-slice-style evaluation of public speech transcripts, more work is needed to unpack the fine details of how scientists communicate their findings and how audiences respond to them. 

\subsubsection*{Summary and Conclusion}
This study demonstrates the power of thin-slicing for evaluating science presentations. Even brief excerpts from the start of presentation talks are sufficient to predict overall quality. This aligns with thin-slicing effects in the nonverbal domain and extend them towards minimal linguistic cues. This approach, particularly when combined with LLMs, offers exciting possibilities for automated feedback and personalized, AI-augmented communication training.  

\section*{Data and Code availability}
Anonymized data and code are available via GitHub at \url {https://github.com/nomcomm/thinslice_ps}.

\bibliographystyle{elsarticle-harv}
\bibliography{main}
\newpage
\section*{Supplementary Methods and Results}
\subsection*{Convergence between Human and LLM Ratings}

\begin{figure}[ht]
  \centering
  \includegraphics[width=0.5\textwidth]{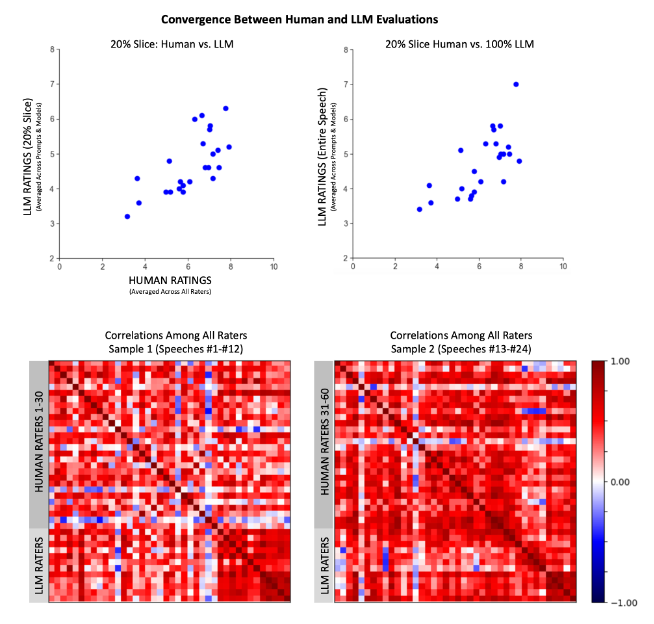}
  \caption*{\textbf{Figure S1.} Convergence between Human and LLM Ratings. Top Row. Group-averaged scores converge (around \textit{r} = 0.7) for both 20\% slices and well as the slice-to-full-speech correlation. Bottom Row: Correlations between Human-to-human raters as well as the different LLM models and prompts (10 last rows) reveal strong positive correlations among all rating sources. LLM-ratings are slightly more consistent/less variable. }
  \label{fig:s1}
\end{figure}

\newpage
\subsection*{Different Prompts Used for LLMs}
Prompt 1: "Here is a transcript from a public presentation on a science/research topic. Please rate the speech quality on a scale from 1 (worst) to 10 (best). Consider factors such as clarity, engagement, and how easy it is to follow. Return only the single rating number as a plain integer, with no other text or characters. Here is the speech text: "

Prompt 2: "You will receive a transcript of a science/research presentation. Rate the overall rhetorical quality on a scale from 1 (worst) to 10 (best), considering clarity, engagement, structure, and delivery. Return only the single rating number as a plain integer, with no other text or characters. Here is the speech text: "

Prompt 3: "Given the following transcript of a science/research presentation, assess its overall speech quality. Focus on aspects such as clarity, engagement, and coherence. Provide only a single numerical rating from 1 (worst) to 10 (best), without any additional text. Here is the speech text: "

Prompt 4: "Imagine you are an expert in public speaking evaluation. Below is a transcript from a science/research presentation. Please rate the effectiveness of the speech on a scale of 1 (worst) to 10 (best) based on clarity, engagement, and ease of understanding. Return only the single rating number as a plain integer, with no other text or characters. Here is the speech text: "

Prompt 5: "Please evaluate the following transcript of a public science/research presentation. Assign a quality rating from 1 (worst) to 10 (best) based on your assessment. Return only a single rating number as a plain integer, with no other text or characters. Here is the speech text: "

\end{document}